\pdfoutput=1

\documentclass[11pt]{article}

\usepackage{ACL2023}

\usepackage{times}
\usepackage{latexsym}
\usepackage{graphicx}

\usepackage{tabularx}
\usepackage{arydshln}

\usepackage{caption}
\usepackage{subcaption}

\usepackage[T1]{fontenc}

\usepackage[utf8]{inputenc}

\usepackage{microtype}

\usepackage{inconsolata}

\title{Extrinsic Factors Affecting the Accuracy of Biomedical NER}

\author{
Zhiyi Li~~~~ 
Shengjie Zhang~~~~
Yujie Song~~~~ 
{Jungyeul Park}\\
Department of Linguistics\\
The University of British Columbia\\
Vancouver, Canada \\
{\tt\{zhiyi,shengjie,yujie\}@student.ubc.ca}~~~ 
{\tt jungyeul@mail.ubc.ca}\\
}

\begin{document}

\maketitle

\begin{abstract}
Biomedical named entity recognition (NER) is a critial task that aims to identify structured information in clinical text, which is often replete with complex, technical terms and a high degree of variability. Accurate and reliable NER can facilitate the extraction and analysis of important biomedical information, which can be used to improve downstream applications including the healthcare system. However, NER in the biomedical domain is challenging due to limited data availability, as the high expertise, time, and expenses are required to annotate its data. In this paper, by using the limited data, we explore various extrinsic factors including the corpus annotation scheme, data augmentation techniques, semi-supervised learning and Brill transformation, to improve the performance of a NER model on a clinical text dataset (i2b2 2012, \citet{sun-rumshisky-uzuner:2013}). 
Our experiments demonstrate that these approaches can significantly improve the model's F1 score from original 73.74 to 77.55. Our findings suggest that considering different extrinsic factors and combining these techniques is a promising approach for improving NER performance in the biomedical domain where the size of data is limited.

\end{abstract}

\section{Introduction}
Identifying entities, such as symptoms and treatment procedures, in the clinical domain can be useful for a variety of applications, including aiding decision making for health care providers in decision support systems \citep{demnerfushman-chapman-mcdonald:2009},
predicting the course of medical conditions in patients, selecting cohorts for conducting retrospective research using electronic health records \citep{nath-lee-lee:2022}.
 monitoring of disease outbreaks, identification of negative effects of medication, and analyzing interactions between medications \citep{zhang:2018}. 
Clinical concept extraction also serves as the foundation for other NLP tasks in the healthcare field, such as identifying relations, question answering and information retrieval \citep{ratinov-roth:2009:CoNLL}.
These tasks can be used to extract and analyze important information from clinical texts, enabling a deeper understanding of patient conditions and treatment options \citep{si-EtAl:2019}.
Obtaining information from clinical reports is difficult because there are limitations on shared data access for privacy reasons and a lack of annotated clinical data for training and evaluating natural language processing (NLP) applications. This requires a high level of expertise in the specific domain \citep{chapman-EtAl:2011,kundeti-EtAl:2016}.
Clinical narrative text is typically written by healthcare professionals for internal communication about a patient's medical history. It includes a variety of progress notes, such as treatment plans, test results, and consultation notes. These documents are often highly specialized and are neither intended for a general audience nor edited for clarity. This can make it challenging to extract useful information from them \citep{leaman-khare-lu:2015}. 

\begin{figure}[t]
     \centering
         \centering
         \includegraphics[width=0.45\textwidth]{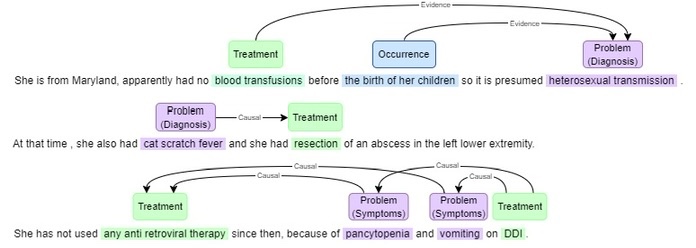}
\caption{{{Example of relations of named entities} \label{shenjie}}}
\end{figure}

The Integrating Biology and the Bedside (i2b2) 2012 dataset is a corpus of clinical text that was created as part of an NLP challenge in 2012 \citep{sun-rumshisky-uzuner:2013}. It has two targeted information extraction tasks: named entity recognition of clinical events and temporal expressions. NER includes annotations for six types of clinical-related events: occurrence, evidential, test, problem, treatment, and clinical departments. By combining the identification of temporal relations with these clinically relevant events, we can reason about the timeline of clinically relevant events for each patient \citep{sun-rumshisky-uzuner:2013}. 
The timeline of clinically relevant events is a key resource for determining diagnoses and treatments and can be used to create automated NLP techniques for computers to extract, speculate, and sequence clinical events in electronic healthcare records. These techniques have the potential to enhance disease tracking, treatment outcome monitoring, and the identification of medication side effects \citep{sun-rumshisky-uzuner:2013}. 
For instance, the problem, test, evidential and occurrence can be combined as evidence for faster diagnosis of the disease and give better treatment plan for the patients. 
{Figure~\ref{shenjie} shows \texttt{treatment} could provide evidence to identify \texttt{problem}} {through biomedical NER}. 

The aim of this paper is to investigate the effect of extrinsic factors on the performance of biomedical NER systems. We conducted a thorough analysis and found that each of these factors consistently improves NER performance. Our findings suggest that modifying the annotation framework can enhance NER performance, and utilizing techniques such as data augmentation semi-supervised annotation and {semi-supervised learning} can add valuable data to improve clinical NER, which is particularly important due to the time and resource-intensive nature of manual annotation in the clinical domain.
{Finally, Brill's transformation-based learning is used to further improve the performance by making rules to correct errors.}

\section{Extrinsic Factors}

\paragraph{Corpus annotation}
The BIO annotation scheme is a widely used method in named entity recognition (NER). It involves labeling each word in the text with a B-tag (beginning), I-tag (inside), or O-tag (outside). The B-tag is used to mark the first word of a named entity, while the I-tag is used to mark subsequent words within the named entity. The O-tag is used to mark words that do not belong to any named entity. 
BIOES (beginning, inside, outside, end, singleton) is a variation of the BIO scheme that includes additional tags for marking the last word of a named entity (E-tag) and single-word named entities (S-tag). 
IO (inside, outside) is another annotation scheme that is similar to BIO, but it only uses I-tags and O-tags. The I-tag is used to mark words that belong to a named entity, while the O-tag is used to mark words that do not belong to any named entity.
{BIOES could potentially outperforms the BIO and IO approaches because it is more granular by including additional tags; these tags could help to better distinguish between the last word of named entity and singleton entities.}

\paragraph{Data augmentation}
Data augmentation is a technique that is commonly used in natural language processing (NLP) to improve the performance of machine learning models. In the context of named entity recognition (NER), data augmentation involves generating additional training examples by modifying or synthesizing existing examples. This can help to improve the generalization of the NER model and make it more robust to variations in language and context \citep{wei-zou-2019-eda}. We used the following methods for data augmentation from \citet{dai-adel-2020-analysis}:
{Label-wise token replacement} (\textsc{lwtr}) uses binomial distribution to randomly decide whether to replace a token of the same label, based on a label-wise token distribution built from a training set.
{Synonym replacement} (\textsc{sr}) is similar to label-wise token replacement but replace the token from the synonyms from WordNet \citep{miller:1995,fellbaum:2006}. 
{Shuffle within segments} (\textsc{sis})  splits the token sequence into segments of the same label and use a binomial distribution to decide whether to shuffle a segment. 
{By doing so, these techniques can help to create more varied and diverse examples. The increasing size and diversity of the training dataset could prevent overfitting and improve robustness and generalizability of the NER model. This is particularly beneficial when the training dataset is small, as generating invalid instances may have a negative impact on the model's performance if the dataset is already large and diverse enough \citep{dai-adel-2020-analysis}.}

\paragraph{Semi supervised learning}

We employ the trained NER model to improve NER results using semi-supervised learning, in which we automatically annotate a large monolingual corpus.
This kind of practice is often called self-training \citep{mcclosky-charniak-johnson:2006:HLT-NAACL06-Main}, self-taught learning \citep{raina-EtAl:2007:ICML}, and lightly-supervised training \citep{schwenk:2002:IWSLT}. 
For semi-supervised learning we introduce the consensus method $\hat{\mathcal{D}}$ \cite{brodley-friedl:1999}.  
We use it by intersection between entity-annotated results using $\hat{\mathcal{D}} = {\mathcal{D}} (\mathcal{M}_{1}) \cap \cdots \cap \mathcal{D} (\mathcal{M}_{n})$ where ${\mathcal{D}}$ is raw text data, $\mathcal{M}_{i}$ is a learning model to annotate raw text data ($1 \leq i \leq n$), and $\hat{\mathcal{D}}$ is filtered annotated data.

\paragraph{Transformation-based learning}
Finally, we applied Brill transformation to improve the performance of the model. Brill transformation is a method that was originally developed for part-of-speech tagging by identifying and correcting errors. It works by applying a series of rules or transformations to the output of an NLP system iteratively, in an attempt to correct errors \citep{brill:1995:CL}. 


\section{Experiments and Results}

We use the neural network architecture of bidirectional LSTM-CNNs for NER proposed in \citet{chiu-nichols-2016-named}. 
This model automatically detects word-level and character-level features using a hybrid bidirectional LSTM and CNN architecture for CRFs, without the need for manual feature engineering. Additionally, the model utilizes state-of-the-art word embeddings, BERT \citep{devlin-EtAl:2019:NAACL}, to handle contextual information. 

After training our data and predicting results using different annotation schemes such as IO or BIOES, we convert the results back to the BIO annotation scheme in order to fairly compare the performance of the different annotation methods. We find that the BIOES annotation scheme outperforms the other two methods, as it achieves a higher F1 score. Therefore, we decide to use the BIOES annotation scheme for our subsequent experiments.

\begin{center}
\begin{tabular}{cccc}
     &  BIO & IO & BIOES\\ \hline
F1 score    & 73.74 & 70.29 & 74.47
\end{tabular}    
\end{center}

To increase the size of our training dataset, we applied data augmentation techniques such as label-wise token replacement, synonym replacement, and shuffle within segments to the original training dataset with BIOES annotation scheme. 
These techniques helped to generate new data that preserved the label information, while also introducing variability and diversity into the dataset. The result of this was an improvement in the F1 score. 


Additionally, we utilized two models to perform semi-supervised learning on the discharge summaries from the \textit{Beth Israel Deaconess Medical Centre}. 
The new dataset is derived from a portion of the i2b2 2010 training dataset (due to IRB restrictions, only a part of the data is available). The original purpose of the i2b2 2010 dataset was to extract medical concepts from patient records \citep{uzuner-EtAl:2011}. 
The new dataset is similar to the 2012 dataset in that both consist of discharge summaries for patients and are intended for the extraction of medical events from healthcare notes. Therefore, the two datasets are similar in style and format while the new dataset adds more variability and diversity. One model was trained on the original dataset with BIOES annotation ($\mathcal{M}_1$), while the other model was trained on the same dataset with the addition of the three data augmentation techniques ($\mathcal{M}_2$). We then used these models to tag the summaries and filtered the resulting entities by intersecting the predictions of the two models. Any sentences that were only tagged with O tags were dropped. This process resulted in further improvement in the results.

For transformation-based learning, we used the model trained after semi-supervised learning as the initial tagger for NLTK's brill tagger. The brill tagger then learned the rules based on half of the original training dataset and was evaluated on the second half of the training data. Since the initial tagger already performed well, we only applied the most confident rules from Brill transformation by setting min acc=0.99 and choosing the best min score based on the evaluation of the second half of the training data. {The final result, with a min score of 5, resulted in an improvement in the F1 score on the testing data to 77.55}

\begin{table*}[th]
    \centering
\begin{tabular}{rcc c c c c} \hline 
&&&& F1 & Size &\\\hline 
original data  & & BIO & $\mathcal{M}$ & 73.74 & 97,759 & $_{\textsc{m}_0}$\\ \hdashline
corpus annotation& ($\mathcal{M}_{ann}$) &BIOES & {\color{white}+} $\mathcal{M}_{ann}$ & 74.47 & 97,759 & $_{\textsc{m}_1}$\\ \hdashline
+ data augumentation &($\mathcal{M}^{daug}$) & BIOES & 
+ $\mathcal{M}_{ann}^{{daug}}$  & 76.60 & 390,417 & $_{\textsc{m}_2}$\\ \hdashline    
+ semi-supervised & ($\mathcal{M}^{semi}$) & BIOES & 
+ $\mathcal{M}_{ann}^{semi} $ & {77.52} & 472,750 & $_{\textsc{m}_3}$\\ \hdashline

+ transformation-based & ($\mathcal{M}^{brill}$) & BIOES & 
+ $\mathcal{M}_{ann}^{brill}$ & \textbf{77.55} & 472,750 & $_{\textsc{m}_4}$\\ \hline

\end{tabular}
    \caption{Results of extrinsic factors and its sizes (the number of tokens) of the training corpus: All predicted results are converted to back the original BIO format to report F1 scores.}
    \label{big-result}
\end{table*}

\section{Discussion}

{For the annotation approaches, the results of the experiments indicate that the BIOES annotation scheme outperforms both the BIO and IO annotation schemes in terms of F1 score across all label categories. Specifically, the BIOES annotation scheme performed better in identifying precise entities and distinguishing boundaries between words. While the BIOES annotation scheme identified fewer phrases (13,283) than the IO and BIO schemes (13,826 and 13,423, resepctively), it identified a higher proportion of correct entities (10,007 vs. 9637 and 9961). This confirms the hypothesis that incorporating additional tags in the annotation process can lead to improved performance in named entity recognition tasks.}

{Our experiments revealed that incorporating data augmentation techniques into the model led to an increase in both the number of identified phrases and the proportion of correct entities. For label categories such as \texttt{clinical dept}, \texttt{problem}, and \texttt{treatment}, where the model identified fewer phrases than the BIOES annotation scheme alone ($_{\textsc{m}_0}$), the model improved the precision score while maintaining a similar recall rate. Conversely, for categories where the model identified more phrases, such as the \texttt{evidential} tag, the precision score decreased from 81.30 to 74.32, while the recall rate increased from 65.04 to 73.95. As a result, the F1 score for the \texttt{evidential} tag category improved from 79.52 to 81.59. The use of data augmentation techniques helped to make the model more generalizable and prevented overfitting, particularly in smaller datasets. While it may increase the number of false positives, the overall improvement in the F1 score suggests that the benefits of data augmentation outweigh this trade-off in the biomedical named entity recognition task.}

{Our experiments revealed that incorporating semi-supervised learning into the model also led to an increase in the amount of data used in training. However, the impact of this approach on the model's performance was distinct from that of data augmentation. The model identified fewer phrases and fewer correct phrases than with either the BIOES annotation scheme alone or data augmentation. Despite this, the overall recall rate decreased from 76.45 to 73.43, while the precision rate increased from 76.75 to 82.09, resulting in an improvement in the overall F1 score. The increase in precision rate was observed across all label categories, while the recall rate remained similar to that of the data augmentation approach for most labels, with the exception of the \texttt{evidential} and \texttt{occurrence} labels. These results suggest that semi-supervised learning may be effective in improving the precision rate for named entity recognition, while data augmentation is better suited for increasing the recall rate. However, further investigation on other datasets is necessary to confirm these findings.}

{In our final set of experiments, we applied Brill transformation to the model and observed an increase in the number of correctly identified phrases for the \texttt{occurrence}, \texttt{test}, and \texttt{treatment} label categories. Specifically, the model found 19 more correct phrases in the \texttt{occurrence} category, 3 more in the \texttt{test} category, and 3 more in the \texttt{treatment} category. Our results suggest that by setting a higher min score for generating Brill rules, the model can identify more accurate rules and improve performance. However, it should be noted that this approach also poses a trade-off, as setting a higher min score can make it more difficult for labels with fewer phrases, such as \texttt{clinical dept} and \texttt{evidential}, to generate rules that pass the threshold. Therefore, it is important to consider the number of phrases for each label category when determining the appropriate minimum score for creating Brill rules.}


\section{Conclusion}
{In conclusion, our research aimed to improve the performance of a biomedical named entity recognition (NER) model by exploring various extrinsic factors, including corpus annotation scheme, data augmentation techniques, semi-supervised learning and Brill transformation. Our experiments on the i2b2 2012 clinical text dataset revealed that these approaches can significantly improve the model's F1 score. We found that the BIOES annotation scheme outperforms both the BIO and IO annotation schemes in terms of F1 score across all label categories. Additionally, incorporating data augmentation techniques and semi-supervised learning into the model can improve the generalizability and prevent overfitting, especially in smaller datasets. The results of our experiments also showed that the use of Brill transformation can increase the number of correctly identified phrases for specific label categories. Our findings suggest that considering different extrinsic factors and combining these techniques is a promising approach for improving NER performance in the biomedical domain, where the size of data is limited. However, more investigation on other datasets is needed to confirm these findings.}


\section*{Acknowledgement}
This material is based upon work partially supported by MITACS \textit{Business Strategy Internship} (IT32060) to Zhiyi Li.


\appendix

\section{Neural Experiment Description}

We use the default setting, as described in the following table:

\begin{center}
\begin{tabular}{c|c} \hline
learning rate &  0.001 \\ 
 epochs & 5 \\
 learning rate decay & 0.005 \\
 batch size & 32 \\\hline
\end{tabular}
\end{center}

The neural network models are run on Google Colab Plus by using high Ram, standard GPU. It takes a few hours to finish training. 
All of the neural models are run by using default setting and no hyperparameters search was used. For Brill transformation, the hyperparameters were discussed in the paper; min acc=0.99 and min score = 2 to 5 was evaluated on the training dataset and picked. 
The neural networks are by single run with random seed = 12
For neural network models, we used \texttt{bert-large-cased} as embeddings and the model structure is \texttt{NerDLApproach} from \texttt{Spark NLP}. For the data augmentation, we used these techniques by modifying the functions from \url{https://github.com/Hironsan/neraug}. And for the Brill transformation, the package is from \texttt{nltk}.



\end{document}